\documentclass[11pt]{article}

\usepackage[preprint]{acl}

\usepackage{times}
\usepackage{latexsym}

\usepackage[T1]{fontenc}

\usepackage[utf8]{inputenc}

\usepackage{microtype}

\usepackage{inconsolata}

\usepackage{graphicx}
\usepackage{tcolorbox}
\usepackage{tabularx}

\usepackage{CJKutf8}
\usepackage{xspace}
\newcommand{\method}{HPO\xspace}
\usepackage{booktabs}

\usepackage{amsmath,amsfonts,bm}

\def\eqref#1{equation~\ref{#1}}

\def\1{\bm{1}}

\def\vf{{\bm{f}}}

\def\vs{{\bm{s}}}

\def\vx{{\bm{x}}}
\def\vy{{\bm{y}}}

\def\mA{{\bm{A}}}

\DeclareMathAlphabet{\mathsfit}{\encodingdefault}{\sfdefault}{m}{sl}
\SetMathAlphabet{\mathsfit}{bold}{\encodingdefault}{\sfdefault}{bx}{n}

\title{Hierarchical Policy Optimization for Simultaneous Translation of Unbounded Speech}

\author{
 \textbf{Siqi Ouyang\textsuperscript{1}},
 \textbf{Shuoyang Ding\textsuperscript{2}},
 \textbf{Oleksii Hrinchuk\textsuperscript{2}},
 \textbf{Vitaly Lavrukhin\textsuperscript{2}},
\\
 \textbf{Brian Yan\textsuperscript{1}},
 \textbf{Boris Ginsburg\textsuperscript{2}},
 \textbf{Lei Li\textsuperscript{1}},
\\
 \textsuperscript{1}Carnegie Mellon University,
 \textsuperscript{2}NVIDIA,
\\
 \texttt{siqiouya@andrew.cmu.edu}
}

\begin{document}
\maketitle

\begin{abstract}
Simultaneous speech translation (SST) generates translations while receiving partial speech input. Recent advances show that large language models (LLMs) can substantially improve SST quality, but at the cost of high computational overhead. To reduce this cost, prior work reformulates SST as a multi-turn dialogue task, enabling full reuse of the LLM’s key–value (KV) cache and eliminating redundant feature recomputation. However, this approach relies on supervised fine-tuning (SFT) data in dialogue form, for which few human annotations exist, and existing synthesis methods cannot guarantee data quality.
In this work, we propose a Hierarchical Policy Optimization (\method) approach that post-train models trained on imperfect SFT data. We introduce a hierarchical reward that balances translation quality and latency objectives. Experiments on English to Chinese/German/Japanese demonstrate improvements of over +7 COMET score and +1.25 MetricX score at a latency of 1.5 seconds. Comprehensive ablation studies further validate the effectiveness of different quality rewards, hierarchical reward formulations, and segmentation strategies. Code can be found here \url{https://github.com/owaski/HPO}. 
\end{abstract}
\section{Introduction}


%
Simultaneous speech translation (SST) generates translations while receiving partial speech input. SST has a wide range of applications, including multilingual conferences, live streaming, and real-time conversations. An SST model must decide, given the currently received speech and the already generated translation, whether to \textit{read} more speech or \textit{write} the next translation token~\citep{ma-etal-2020-simulmt,ren-etal-2020-simulspeech}. 

Recent state-of-the-art approaches train SST models using synthesized read–write trajectory~\citep{wang-etal-2025-conversational,ouyang-etal-2025-infinisst,fu2025efficientadaptivesimultaneousspeech,cheng2025seedliveinterpret20endtoend}. This paradigm enables inference-efficient architectures that scale beyond short speech utterances to unbounded speech that spans minutes or even hours. Formulating SST as a multi-turn dialogue, where speech input and translation output interleave, allows the model to reuse key–value (KV) caches across both modalities. This reuse eliminates redundant feature recomputation during inference and ensures efficient handling of long-context streaming speech.

\begin{CJK*}{UTF8}{gkai}
However, existing methods of synthesizing trajectories have notable limitations. There are two main approaches. The first relies on word-alignment tools~\citep{wang-etal-2025-conversational,ouyang-etal-2025-infinisst}, which account for word reordering between source and target but ignore the future context needed for accurate translation timing.
The second approach uses large language models (LLMs) to mimic human interpreters~\citep{makinae-etal-2024-simul,cheng2024achievinghumanparityendtoend,fu2025efficientadaptivesimultaneousspeech}. These methods segment the source transcript into smaller units, each deemed sufficient for translation. However, such segmentation is often unstable and provides no guarantee of producing valid read-write trajectories.  

\end{CJK*}

In this paper, we propose Hierarchical Policy Optimization (\method), a post-training approach designed to correct erroneous behaviors arising from imperfections in synthesized training trajectories.
We adapt Group Relative Policy Optimization (GRPO)~\citep{shao2024deepseekmathpushinglimitsmathematical} to jointly optimize for translation quality and latency. Since the latency reward is inherently easier to optimize, which simply encourages the model to translate earlier regardless of correctness~\citep{xu-etal-2025-seqpo}, we introduce a hierarchical reward structure to prevent over-optimization toward latency. Specifically, if the translation quality does not exceed a predefined threshold, the latency reward is set to its minimum, ensuring that the model prioritizes accuracy before speed. To further stabilize training, we apply group normalization separately to the quality and latency rewards before combining them into a single overall reward signal.
Experiments on the ACL 60/60 development set~\citep{salesky-etal-2023-evaluating} and RealSI~\citep{cheng2024achievinghumanparityendtoend} demonstrate that \method consistently improves translation quality across a wide range of latency levels. For instance, at an average latency of 1.5 seconds, \method achieves a +7 COMET, +1.25 MetricX and +4 BLEURT improvement over the strongest baseline.

\section{Related Works}

\paragraph{SST with LLM}

Recent studies have demonstrated that the translation quality of SST can be substantially improved by adopting LLM as the backbone~\cite{ahmad-etal-2024-findings}. \citet{koshkin-etal-2024-transllama} showed that an LLM can be adapted for SST by finetuning on a small set of synthetic translation trajectories. \citet{ouyang-etal-2025-infinisst} further extended this idea with an interleaving architecture that enables efficient inference over unbounded speech input. Similarly, \citet{fu2025efficientadaptivesimultaneousspeech} explored similar architectures but focused on alternative training strategies and data synthesis pipelines. \citet{guo2025streamuniachievingstreamingspeech} proposed a unified model for both streaming transcription and translation, along with a truncation mechanism that prunes historical speech and text based on automatically transcribed inputs.
However, most prior methods rely on either heuristic policies such as wait-$k$~\cite{ma-etal-2019-stacl} or synthetic translation trajectories without quality guarantees. In contrast, \method introduces reinforcement learning to further refine models trained on synthetic data, improving their robustness and alignment with human preferences. 

\paragraph{Reinforcement Learning for SST}

Previous work has primarily applied reinforcement learning (RL) to optimize policies for simultaneous text translation~\cite{grissom-ii-etal-2014-dont,gu-etal-2017-learning,alinejad-etal-2018-prediction,ive-etal-2021-exploiting,wang2022simultaneous,xu-etal-2025-seqpo}. The core idea is to treat the incoming source stream as an environment, where the agent observes partial input and learns a policy to decide when to \emph{read} or \emph{write} based on the current state and past translations. \citet{gu-etal-2017-learning} investigated different latency objectives and combined quality and latency rewards through simple additive weighting. More recently, \citet{xu-etal-2025-seqpo} proposed SeqPO-SiMT, which normalizes quality and latency rewards separately and truncates the latency component before combining them to mitigate optimization instability caused by the scale difference of quality and latency rewards.
However, these studies focus exclusively on text-based translation and generally rely on encoder–decoder Transformers rather than LLMs. Only SeqPO-SiMT incorporates an LLM backbone, and none of the existing methods support unbounded speech input. In contrast, \method extends RL optimization to SST, enabling efficient and robust inference over continuous audio streams.

\begin{figure}[t]
    \centering
    \includegraphics[width=\linewidth]{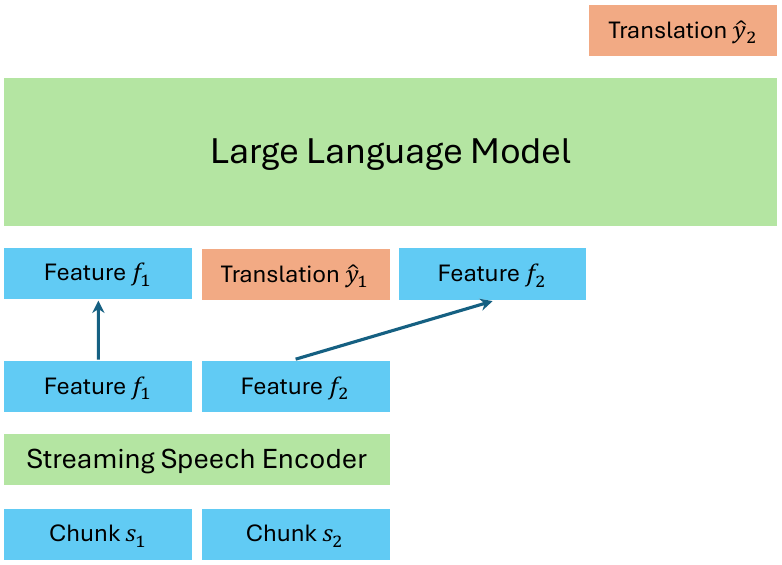}
    \caption{Model architecture of \method. Speech chunks are encoded by a streaming speech encoder into contextualized features. The large language model then takes interleaved speech features and prior translations as input to decode the next partial translation.}
    \label{fig:arch}
    \vspace{-0.5cm}
\end{figure}

\begin{figure*}[t]
    \centering
    \includegraphics[width=\linewidth]{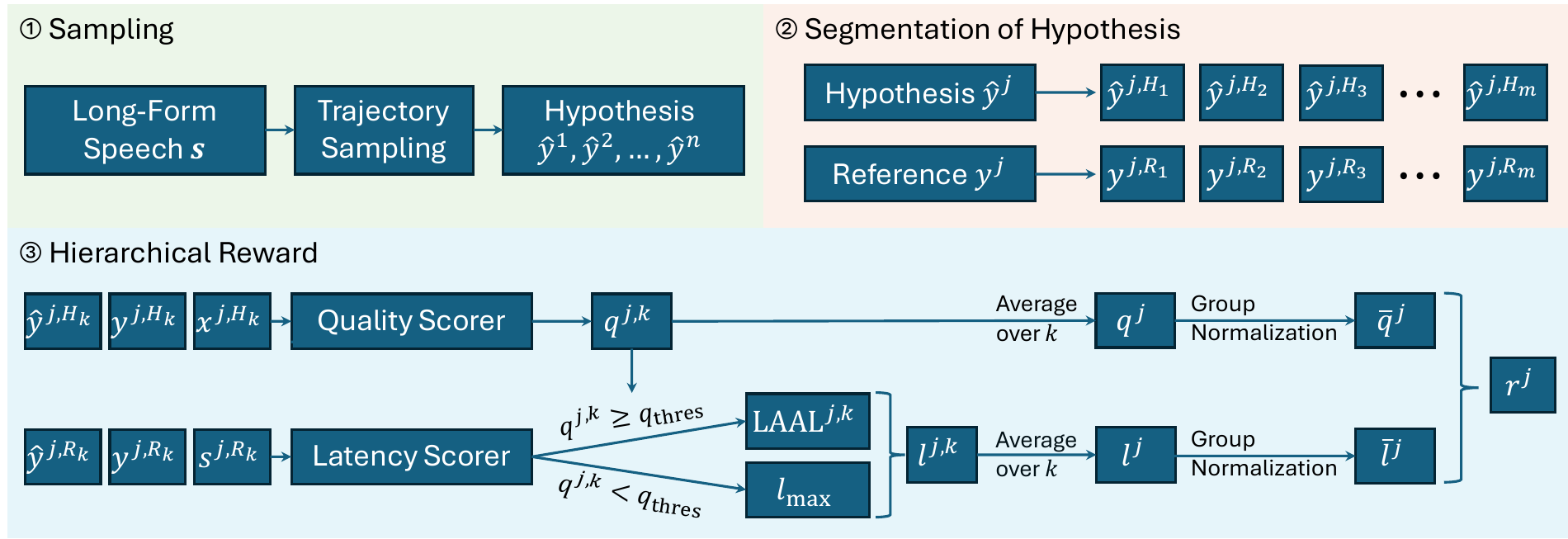}
    \caption{Overview of Hierarchical Policy Optimization. Given a long-form speech input $\vs$, we sample $n$ hypotheses $\hat{\vy}^1,\dots,\hat{\vy}^n$. Each hypothesis $\hat{\vy}^j$ is segmented into sentences $\hat{\vy}^{j,H_1},\dots,\hat{\vy}^{j,H_m}$ and aligned with the corresponding reference translation sentences $\vy^{j,R_1},\dots,\vy^{j,R_m}$. For each aligned sentence pair $(H_k, R_k)$, we compute a quality score $q^{j,k}$ and a latency score $l^{j,k}$. Latency is optimized only when the corresponding quality score exceeds the predefined threshold $q_\text{thres}$. Finally, we average the quality and latency scores across all sentences of each hypothesis $\hat{\vy}^j$, apply group normalization to both components, and sum them to obtain the final reward.}
    \label{fig:hpo}
    \vspace{-0.5cm}
\end{figure*}

\section{Preliminaries}

\subsection{Problem Formulation}

Let $\vs = (s_1, s_2, \dots, s_T) \in \mathbb{R}^T$ denote the source speech waveform. At each step $i$, the environment emits a fixed-duration speech chunk
\[
\vs_{i} = \left(s_{i \cdot c + 1}, \dots, s_{(i+1) \cdot c}\right),
\]
where $c$ is the chunk length. The policy function $\pi_\theta(\vs_{1:i}, \hat{\vy}_{1:i-1})$ then generates a partial translation $\hat{\vy}_i$ conditioned on all past speech chunks $\vs_{1:i}$ and previously generated translations $\hat{\vy}_{1:i-1}$. The output $\hat{\vy}_i = (\hat{y}^i_1, \dots, \hat{y}^i_{|\hat{\vy}_i|})$ may contain a variable number of tokens. When $|\hat{\vy}_i| = 0$, the policy chooses to wait for additional speech input before producing any translation. When $|\hat{\vy}_i| > 0$, the policy outputs a partial translation. All tokens in $\hat{\vy}_i$ are assigned the same delay of $i \cdot c$, reflecting the time elapsed when the chunk $\vs_i$ is observed. Translation quality and latency are then evaluated with respect to the full source waveform $\vs$, the aggregated translation hypothesis $\hat{\vy}$, the ground-truth translation $\vy$, the source transcript $\vx$ and the delay associated with each generated token.

\subsection{Model Architecture}
\label{sec:arch}

We adopt InfiniSST as the architecture since it achieves the best translation quality in the IWSLT 2025 low-latency track ~\citep{ouyang-etal-2025-cmus, agostinelli-etal-2025-findings}.  

As shown in Figure~\ref{fig:arch}, the system consists of two components: a streaming speech encoder and an LLM translator. 
The streaming speech encoder incrementally encodes each new speech chunk while reusing the KV cache of prior chunks:
\[
    \vf_i = \text{Encode}\big(\vs_{i} \;\big|\; \vs_{1 : i - 1}\big).
\]

The encoded features are fed into the LLM, which decodes the corresponding translation given interleaved speech and text input
\[
    \hat{\vy}_i \sim \text{LLM}\left(\cdot \;\big|\; \vf_1, \hat{\vy}_1, \vf_2,\hat{\vy}_2, \dots, \vf_{i-1}, \hat{\vy}_{i-1}, \vf_i\right).
\]
Decoding stops when the LLM outputs the special token $\langle$\texttt{EOT}$\rangle$. If the LLM immediately produces $\langle$\texttt{EOT}$\rangle$, the translation for that chunk is empty. 

This architecture eliminates feature recomputation and fully reuses the KV cache for both prior speech features and previously decoded translations, making it computationally efficient in practice. For unbounded speech streams, i.e., continuous speech without duration limits, we apply a sliding-window mechanism for the speech encoder and the Attention Sink technique~\citep{xiao2024efficient} for the LLM.  

\subsection{Data Synthesis}
\label{sec:data_synth}

We follow the practice of \citet{ouyang-etal-2025-infinisst} to synthesize the interleaving data. To simulate long-form speech and capture phenomena such as silence, laughter, and background noise, we construct training samples from extended speech recordings spanning minutes to hours. Each recording is divided into fixed-length segments of up to 67.2 seconds (corresponding to 60 chunks of 1.12 seconds each). Speech is first aligned with transcripts using timestamps obtained from an ASR model, after which transcript–translation alignments are extracted using SimAlign~\cite{jalili-sabet-etal-2020-simalign}. These alignments are then used to construct the interleaving speech–text format required for supervised finetuning (SFT). Full details of the synthesis procedure are provided in Appendix~\ref{apdx:data_synth}.

\section{Hierarchical Policy Optimization}

Figure \ref{fig:hpo} shows the overview of \method. We adapt Group Relative Policy Optimization (GRPO)~\citep{shao2024deepseekmathpushinglimitsmathematical} for post-training the model that is trained on synthesized interleaving data. Following GRPO, for each long-form speech segment $\vs$, we sample multiple translation hypotheses $\hat{\vy}^1, \hat{\vy}^2, \dots, \hat{\vy}^n$ and evaluate them in terms of both translation quality and latency. Since each hypothesis $\hat{\vy}^j$ may contain multiple sentences, we first segment it into sentences and align them with the corresponding reference translation sentences of $\vs$. We then compute a hierarchical reward that jointly accounts for quality and latency, and use this reward to optimize the model parameters.

\subsection{Segmentation}

A common approach for segmenting hypotheses into sentences aligned with reference sentences is mwersegmenter~\citep{matusov-etal-2005-evaluating}, which was originally designed to work with surface-level translation quality metrics on document translation evaluation. However, we observe that it often introduces segmentation errors, such as splitting within a sentence.  
To mitigate this, we adopt SEGALE~\citep{wang2025lcme}, which extends sentence-level machine translation evaluation metrics to document-level evaluation. It employs an off-the-shelf sentence segmenter, such as spaCy~\footnote{\url{https://spacy.io/}}, to split long-form texts into sentences, and then uses an embedding-based aligner~\citep{thompson-koehn-2019-vecalign} combined with an adaptive search procedure to robustly align hypothesis and reference sentences while identifying under- and over-translation errors.  

Formally, each hypothesis \(\hat{\vy}^j\) is segmented by spaCy\footnote{We use the transformer backend.} into sentences \(\hat{\vy}^{j,1}, \dots, \hat{\vy}^{j,m_h}\), and the pre-segmented reference translation sentences are denoted as \(\vy^{j,1}, \dots, \vy^{j,m_r}\). SEGALE produces an alignment $\mA = (A_1, A_2, \dots, A_m)$, where each $A_k = \{H_k, R_k\}$ consists of a set $H_k$ of hypothesis sentence indices and a set $R_k$ of aligned reference indices. There could be null alignments. If $R_k = \phi$, the alignment indicates an over-translation in the hypothesis, while if $H_k = \phi$, it indicates an under-translation.  Given such alignment, we then calculate the hierarchical reward.

\subsection{Hierarchical Reward}

Given the aligned tuple (hypothesis $\hat{\vy}^{j, H_k}$, reference $\vy^{j,R_k}$, source transcript $\vx^{j,R_k}$, source speech $\vs^{R_k}$), we first compute the quality score $q^{j,k}$ and latency score $l^{j,k}$ separately. The quality score can be estimated using existing translation metrics such as COMET~\citep{guerreiro-etal-2024-xcomet} and MetricX~\citep{juraska-etal-2024-metricx}. The latency score is estimated using length adaptive average lagging (LAAL)~\cite{papi-etal-2022-generation}. Specifically, we use the start time of the source speech segment $\vs^{R_k}$ to offset the start time of the hypothesis sentence $\hat{\vy}^{j,H_k}$, assuming the source speech starts at time 0. We then compute LAAL of this tuple given the source speech duration, the reference length, and the hypothesis delay.

If either \(H_k=\phi\) or \(R_k=\phi\), indicating over/less-translation, we assign the worst possible quality and latency scores to penalize such behavior. For instance, in MetricX with scale from \(-25\) to \(0\), the worst score is \(-25\); for latency, we set \(l_{\max}=10\) seconds, since empirically most translation trajectory will have latency smaller than this. 

The latency score is easier to optimize compared to quality score as it simply needs the model to generate the translation early without assuring its translation quality. Thus, we set the latency score to be its maximum $l_{\max}$ if the quality score is below a certain threshold $q_\text{thres}$,
\begin{align}
    l^{j,k} = \begin{cases}
        \text{LAAL}^{j,k} & q^{j,k} \geq q_\text{thres} \\
        l_{\max} & q^{j,k} < q_\text{thres}
    \end{cases}
\end{align}
This design effectively mitigates over-optimization toward latency which degrades translation quality. For the threshold $q_\text{thres}$, for example for MetricX we set it to be -5.

Then we average the quality and latency scores of sample $j$, 
\begin{align}
    q^j = \frac{1}{m}\sum_{k=1}^m q^{j,k},\quad l^{j} = \frac{1}{m}\sum_{k=1}^m l^{j,k}, 
\end{align}

Finally, we apply group normalization to the quality and latency scores separately and then add them together,
\begin{align}
    \bar{q}^j &= \frac{q^j - \text{mean}(q^1,\cdots,q^n)}{\text{std}(q^1,\cdots,q^n)} \\
    \bar{l}^j &= \frac{l^j - \text{mean}(l^1,\cdots,l^n)}{\text{std}(q^1, \cdots,q^n)} \\
    r^j &= \bar{q}^j - \lambda  \cdot \bar{l}^j,
\end{align}
where $\lambda$ controls the weight of latency reward. 

\subsection{Optimization}

The overall training objective of HPO is defined as
\begin{equation}
\label{eq:hpo-objective}
J(\theta) =
\mathop{\mathbb{E}}\limits_{\substack{
    \vs \sim p_\text{data} \\
    \hat{\vy}^1, \dots, \hat{\vy}^n \sim \pi_\theta(\vs)
}}
\left[ \frac{1}{n} \sum_{j=1}^n \frac{1}{|\hat{\vy}^j|}
       \sum_{t=1}^{|\hat{\vy}^j|} R^j_t \right].
\end{equation}
where $p_\text{data}$ is the training data distribution and $R^j_t$ is token-level reward which is computed as
\begin{equation}
R^j_t = 
\frac{\pi_\theta}{\pi_{\theta_\text{old}}}
\left[
    C_t^j - \beta \,
    \mathcal{D}_\text{KL}^\text{on-policy}
\right],
\end{equation}
where $\tfrac{\pi_\theta}{\pi_{\theta_\text{old}}}$ is the importance sampling ratio.  
The clipped reward $C_t^j$ follows GRPO clipping:
\begin{equation}
C_t^j =
\min \left[
    \frac{\pi_\theta}{\pi_{\theta_\text{old}}} r^j,
    \ \text{clip}\!\left(
        \frac{\pi_\theta}{\pi_{\theta_\text{old}}},
        1 - \varepsilon,\, 1 + \varepsilon
    \right) r^j
\right].
\end{equation}
Finally, the on-policy KL divergence is approximated as
\begin{equation}
\mathcal{D}_\text{KL}^\text{on-policy} =
\frac{\pi_\theta}{\pi_{\theta_\text{old}}}
\left(
    \frac{\pi_\text{ref}}{\pi_\theta}
    - \log \frac{\pi_\text{ref}}{\pi_\theta}
    - 1
\right).
\end{equation}
The importance sampling term is used to stabilize training and prevent divergence from $\pi_{\theta_\text{old}}$ during the mini-batch update. 

\begin{figure*}[t]
    \centering
    \includegraphics[width=\linewidth]{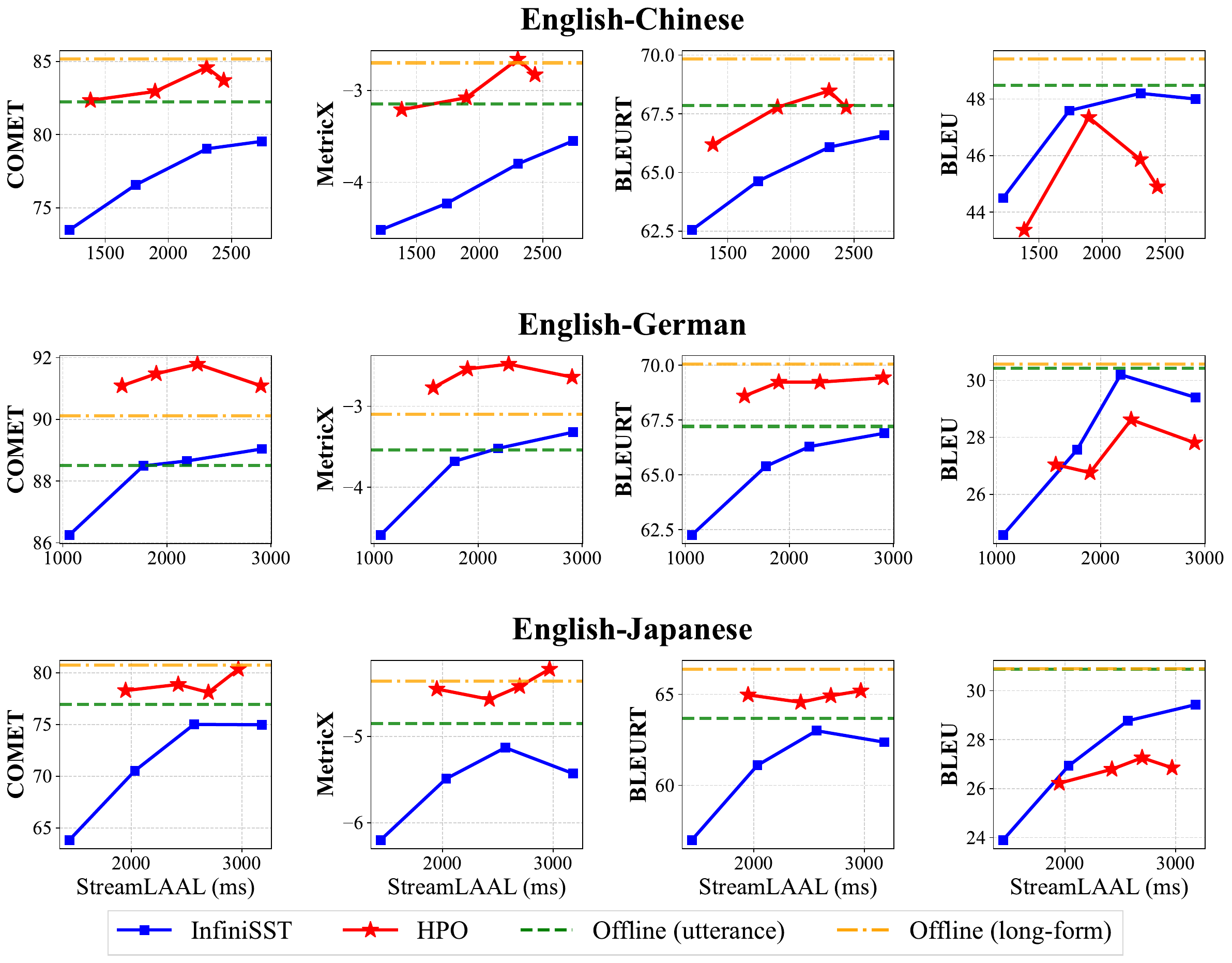}
    \caption{Evaluation results on the ACL 60/60 dev set. Each row corresponds to a language direction (En–Zh/De/Ja), and each column corresponds to a translation quality metric (COMET, MetricX, BLEURT, and BLEU). The Y-axis indicates the quality score and the X-axis indicates latency measured with StreamLAAL. \method achieves consistently higher translation quality than the strong InfiniSST baseline at comparable latency in three out of four metrics, and even surpasses the offline translation model in overall quality.}
    \label{fig:main_acl6060}
    \vspace{-0.5cm}
\end{figure*}

\begin{figure*}[t]
    \centering
    \includegraphics[width=\linewidth]{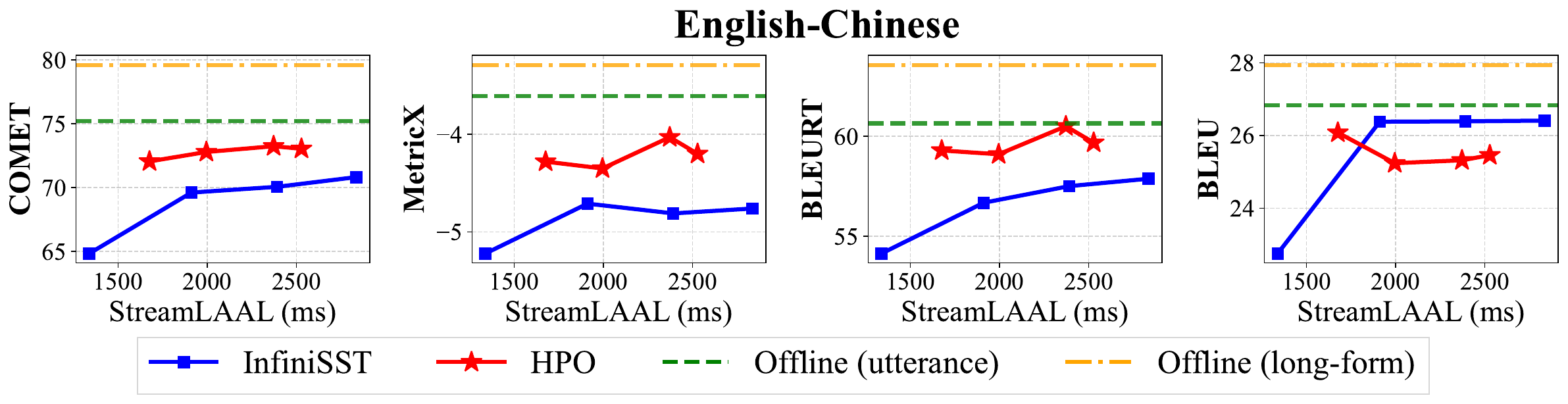}
    \caption{Evaluation results on the RealSI En-Zh test set. Each column corresponds to a translation quality metric (COMET, MetricX, BLEURT, and BLEU). The Y-axis indicates the quality score and the X-axis indicates latency measured with StreamLAAL. \method achieves consistently higher translation quality than the strong InfiniSST baseline at comparable latency in three out of four metrics.}
    \label{fig:main_realsi}
    \vspace{-0.2cm}
\end{figure*}

\begin{figure*}[t]
    \centering
    \includegraphics[width=\linewidth]{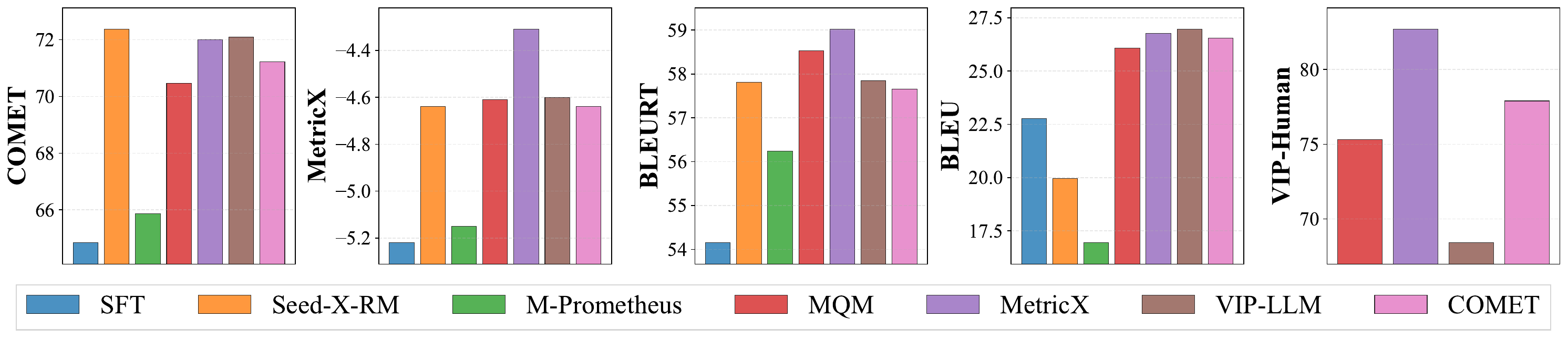}
    \caption{We train \method using six different quality reward functions (Seed-X-RM, M-Prometheus, MQM, MetricX, VIP-LLM, and COMET) and cross-validate their performance across four evaluation metrics (left four figures). 
We further assess four of these reward functions (MQM, MetricX, VIP-LLM, and COMET) through human evaluation (rightmost figure). 
MetricX is the only reward function that consistently achieves competitive performance across all automatic metrics and human judgments.}
    \label{fig:reward}
    \vspace{-0.5cm}
\end{figure*}

\begin{figure}[t]
    \centering
    \includegraphics[width=0.7\linewidth]{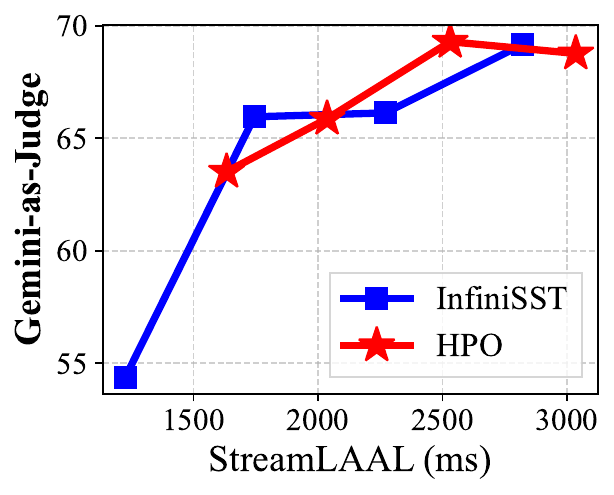}
    \caption{Evaluation results on the ACL 60/60 En-Zh dev set. The Y-axis indicates the Gemini-as-Judge score and the X-axis indicates latency measured with StreamLAAL.}
    \label{fig:gemini_eval}
    \vspace{-0.5cm}
\end{figure}

\section{Experiment Setup}

\subsection{Dataset}

\paragraph{Training} 
Existing speech translation datasets consist mainly of short utterances, e.g., CoVoST2~\citep{wang21s_interspeech}, while others provide long-form speech pre-segmented into single sentences, e.g., MuST-C~\citep{di-gangi-etal-2019-must}, which is no longer distributed due to licensing issues. To train a simultaneous speech translation model capable of handling beyond single-utterance input, we directly construct a long-form dataset derived from YODAS~\citep{10389689}, a large-scale collection of over 500k hours of multilingual YouTube speech. From YODAS, we select a 5k-hour subset of English speech (en000) and build (speech, transcript, translation) triplets together with synthetic interleaving trajectories. More details can be found in Appendix \ref{apdx:data_synth}. 

\paragraph{Evaluation} We evaluate our model and baselines on two datasets. The first is the ACL 60/60 dev set~\citep{salesky-etal-2023-evaluating}, which consists of five academic talks on ACL papers, each lasting 10–20 minutes and translated into 10 languages including Chinese, German, and Japanese. This dataset was also adopted in the recent IWSLT competition~\citep{agostinelli-etal-2025-findings}. The second dataset is RealSI~\citep{cheng2024towards}, which contains 10 talks covering diverse topics. Compared to academic talks, these recordings are more spontaneous and therefore more representative of real-life speech. Each talk lasts on average about 5 minutes, and the dataset is available only for English–Chinese translation.

\subsection{Evaluation Metric}

\paragraph{Latency} Following the practice of the IWSLT 2025, we adopt StreamLAAL~\citep{papi-etal-2024-streamatt} for latency evaluation. We switch the segmenter from mwersegmenter to SEGALE for hypothesis segmentation, which provides more accurate segmentations. For null alignments, we assign the latency of 10 seconds as penalty. 

\paragraph{Translation Quality}
We evaluate the aligned hypothesis sentences and reference sentences using five translation quality metrics: BLEU~\citep{papineni-etal-2002-bleu}, BLEURT-20~\citep{pu-etal-2021-learning}\footnote{\url{https://huggingface.co/lucadiliello/BLEURT-20}}, COMET~\citep{guerreiro-etal-2024-xcomet,rei-etal-2023-scaling}\footnote{\url{https://huggingface.co/Unbabel/XCOMET-XXL}}, MetricX~\citep{juraska-etal-2024-metricx}\footnote{\url{https://huggingface.co/google/metricx-24-hybrid-xxl-v2p6-bfloat16}}, and LLM-as-Judge with Gemini-3.1-Pro-Preview high-thinking effort, following the prompt in Appendix A of Findings of the WMT25 Shared Task on Automated Translation Evaluation Systems~\citep{lavie-etal-2025-findings}. For null alignments, we assign the worst quality scores (e.g., -25 for MetricX, 0 for COMET). 

\subsection{Model Configuration}

We use Qwen3-4B-Instruct-2507\footnote{\url{https://huggingface.co/Qwen/Qwen3-4B-Instruct-2507}}
 as the base LLM, which supports over 100 languages. As the speech encoder, we adopt the cache-aware Fast Conformer~\citep{10446861} from a streaming ASR model\footnote{\url{https://huggingface.co/nvidia/stt_en_fastconformer_hybrid_large_streaming_multi}}
 trained on several thousand hours of English speech. To bridge the speech and text modalities, we append two additional Fast Conformer layers after the encoder, configured identically to the original encoder layers, serving as a lightweight modality adapter.

Details of SFT with synthetic trajectories are provided in Appendix~\ref{apdx:sft_train}. For \method, we reuse the same synthetic dataset. At each step, we sample 32 speech segments, and for each segment, generate 16 translation trajectories using top-$p=0.999$ and top-$k=10000$ sampling. The mini-batch size is set to 128. The KL penalty weight is 0.01, and the reward clipping ratio $\varepsilon$ is 0.2. We apply gradient norm clipping at 1.0 and use the Adam optimizer with a learning rate of $1\times10^{-6}$, weight decay of 0.01, and $(\beta_1, \beta_2) = (0.9, 0.999)$. MetricX serves as the default quality reward model, with threshold $q_\text{thres} = -5$, and the latency reward weight is $\lambda = 0.5$. We train the model with NeMo-RL~\footnote{\url{https://github.com/NVIDIA-NeMo/RL}} for up to 700 steps and select the checkpoint with the highest validation quality reward. A single HPO training run takes about 20 hours for 500 steps on three 8xH100 nodes, where one node is dedicated to reward computation, while the other two are used for colocated model training, inference, and rollout generation.

During inference, we use beam search with a beam width of 4 and a repetition penalty of 1.0. To enable streaming inference over long-form speech, we preserve the KV cache of the first 400 tokens and maintains a sliding window of 2000 tokens for subsequent decoding. The KV caches are taken before applying rotary embeddings~\citep{10.1016/j.neucom.2023.127063}; we then concatenate the two cache sets and re-apply the rotary embeddings.


\subsection{Baseline}

\paragraph{Offline ST}~\citet{zhang2023tuninglargelanguagemodel} translates using the full speech context rather than incrementally, serving as an approximate upper bound on translation quality. We train the offline ST model on the same long-form speech data as \method, as well as on utterance-level data extracted from the long-form segments. Since test speeches are too long to fit entirely into the model, we implement two inference modes: (i) utterance-level inference, where the model is given only pre-segmented speech utterances, and (ii) long-form inference, where the model processes up to 67.2 seconds of speech at a time while conditioning on its own translations of prior utterances. Prior work~\citep{ouyang-etal-2025-infinisst, papi-etal-2024-streamatt} has considered only utterance-level offline ST. We include long-form inference for a fairer comparison, since the SST models operates directly on long-form speech.

\paragraph{InfiniSST (SFT)}~\citet{ouyang-etal-2025-infinisst} is the state-of-the-art SST model for unbounded speech, and achieved the best translation quality in the low-latency track at IWSLT 2025~\citep{ouyang-etal-2025-cmus,agostinelli-etal-2025-findings}. InfiniSST treats SST as a multi-turn dialogue and trains the model on synthetic trajectories. We use it as the initialization for \method.

\begin{table*}[t]
    \centering
    \begin{tabular}{l|ccccc}\toprule
        \textbf{Method} & \textbf{StreamLAAL} & \textbf{COMET} & \textbf{MetricX} & \textbf{BLEURT} & \textbf{BLEU} \\\midrule
        SFT & \textbf{1216} & 0.7348 & -4.52 & 0.6255 & \textbf{44.5}\\
        Normalize & 1555 & 0.7977 & -3.41 & 0.6417 & 41.11 \\
        Normalize + Truncation (SeqPO) & 1805 & 0.8058 & -3.39 & 0.6508 & 42.18 \\
        Normalize + Hierarchical-Doc & 1544 & \underline{0.8157} & \underline{-3.27} & \underline{0.6517} & 42.78 \\
        Normalize + Hierarchical-Sent (\method) & \underline{1383} & \textbf{0.8234} & \textbf{-3.21} & \textbf{0.6619} & \underline{43.37} \\
        \bottomrule
    \end{tabular}
    \caption{Ablation on hierarchical reward. \method achieves the overall best translation quality in the latency region.}
    \label{tab:hier}
\end{table*}

\begin{CJK*}{UTF8}{gkai}
\begin{table*}[t]
\centering
\resizebox{\linewidth}{!}{
\begin{tabular}{l|l|l|l|l}
\toprule
\textbf{Source} & \textbf{Reference} & \textbf{Hypothesis} & \textbf{Hypothesis (En Translation)} & \textbf{MetricX} \\ \midrule
\begin{tabular}[c]{@{}l@{}}There is a technique \\ element too.\end{tabular} & 也有一个技术要素。 & \begin{tabular}[c]{@{}l@{}}是的，我知道了。\\ 谢谢。\end{tabular} & \begin{tabular}[c]{@{}l@{}}Yes, I understand.\\Thank you.\end{tabular} &  -0.53 \\ \hline
\begin{tabular}[c]{@{}l@{}}So the Scrum Master \\ will really, really help \\ the product owner on \\ these two fronts.\end{tabular} & \begin{tabular}[c]{@{}l@{}}因此，Scrum Master\\ 将在这两个方面真正、\\ 非常地帮助产品负责人。\end{tabular} & \begin{tabular}[c]{@{}l@{}}好的，我明白。没问题，\\ 我理解了。谢谢，我懂了。\\ 是的，我明白你的意思。\\ 好的，我理解了。\end{tabular} & \begin{tabular}[c]{@{}l@{}}Okay, I understand. No problem, \\ I’ve got it. I understand. Thanks, I got it.  \\ Yes, I understand what you mean. \\ Okay, I understand. \end{tabular}  &  -0.83 \\ \bottomrule
\end{tabular}
}
\caption{A example of gibberish hypothesis segmented by mwersegmenter and achieves near perfect MetricX scores. }
\label{tab:mwer}
\vspace{-0.4cm}
\end{table*}
\end{CJK*}

\section{Results and Analysis}

\subsection{\method achieves the best quality-latency trade-off}

The evaluation results of \method and the baselines are shown in Figure~\ref{fig:main_acl6060} and Figure~\ref{fig:main_realsi}. \method achieves the best trade-off between translation quality and latency in three out of four metrics (COMET, MetricX, and BLEURT) across all three language directions. At a latency of around 1.5 seconds, \method improves COMET by up to 7 points, MetricX by up to 1.25 points, and BLEURT by up to 4 points. Interestingly, \method is competitive with utterance-level offline ST in terms of translation quality and, in some cases, even surpasses long-form offline ST. This highlights the effectiveness of \method. Note that while offline ST could also be optimized with standard RL methods such as GRPO, we did not include these experiments due to computational budget constraints.

BLEU is the only exception across four metrics. This discrepancy raises concerns about possible reward hacking, since MetricX that HPO uses during training is a neural reward. To further examine this issue, we additionally evaluate the En--Zh direction with Gemini, as shown in Figure~\ref{fig:gemini_eval}. The Gemini evaluation supports the concern that optimizing with existing neural rewards like MetricX or COMET may lead to reward hacking. Future work is needed to develop more reliable quality reward. 

\subsection{Ablations}

\paragraph{Quality Reward}

We next evaluate how different quality reward functions affect model performance. In this experiment, we optimize models solely with the quality reward and then cross-validate each model using multiple quality metrics. In total, we consider six reward signals (Seed-X-RM, M-Prometheus, MQM, MetricX, VIP-LLM and COMET) and four quality metrics (COMET, MetricX, BLEURT, BLEU). Seed-X-RM\footnote{\url{https://huggingface.co/ByteDance-Seed/Seed-X-RM-7B}}
 is a reward model proposed by \citet{cheng2025seedxbuildingstrongmultilingual}, while M-Prometheus~\citep{pombal2025mprometheus}\footnote{\url{https://huggingface.co/Unbabel/M-Prometheus-14B}}
 is a multilingual LLM judge. MQM is a GEMBA-MQM–style reward~\citep{kocmi-federmann-2023-gemba}, where we query an instruction model\footnote{\url{https://huggingface.co/Qwen/Qwen3-30B-A3B-Instruct-2507-FP8}}
 eight times and average the scores. Finally, VIP-LLM follows the VIP protocol of \citet{cheng2024towards}, simulating human evaluation using a thinking model\footnote{\url{https://huggingface.co/Qwen/Qwen3-30B-A3B-Thinking-2507-FP8}} (see Appendix~\ref{apdx:prompt_vip} for the prompt). We query it four times and take the majority vote as the final reward.

Evaluation results on RealSI are shown in Figure~\ref{fig:reward}. MQM, VIP-LLM, MetricX, and COMET rewards perform comparably, with MetricX slightly outperforming the other three across most metrics. To further compare MQM, VIP-LLM, MetricX, and COMET rewards, we conduct human evaluation following the VIP protocol~\citep{cheng2024towards} (see Appendix~\ref{apdx:human_eval}). As shown in the rightmost figure of Figure~\ref{fig:reward}, MetricX aligns best with human judgments. We therefore adopt MetricX as the default reward function in all subsequent experiments across language directions.  

\begin{figure*}[t]
    \centering
    \includegraphics[width=\linewidth]{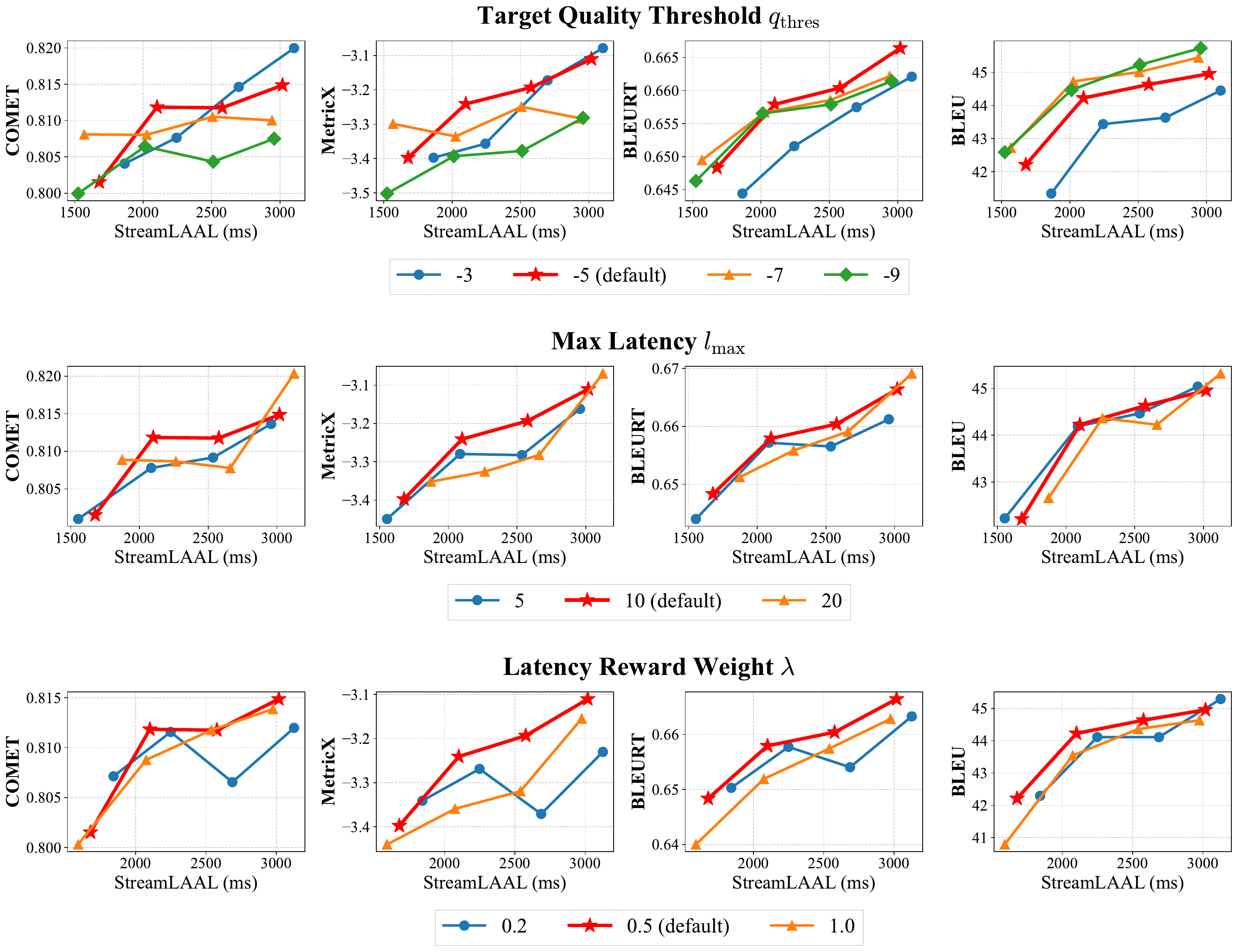}
    \caption{Sensitivity of \method to hyperparameters on the ACL 60/60 dev set. Each row varies one hyperparameter (Target Quality Threshold, Max Latency, or Latency Reward Weight), and each column corresponds to a translation quality metric (COMET, MetricX, BLEURT, or BLEU). The x-axis shows latency measured by StreamLAAL, and the y-axis shows translation quality. Overall, the default hyperparameter setting performs best across configurations.}
    \label{fig:ablation_hyperparam_acl6060}
    \vspace{-0.4cm}
\end{figure*}

\paragraph{Hierarchical Reward}

We evaluate how effective is hierarchical reward. We conduct experiment on different choices of reward combination: \textit{Normalize} means only do normalization for quality and latency separately and add them together. \textit{Normalize + Truncation} is the method used by \citet{xu-etal-2025-seqpo}, which truncates the minimum value of the normalized latency\footnote{We find that truncate latency by the chunk size as in SeqPO is quite unstable during training for our model and we find that truncate by $\frac{\text{chunk size}}{3}$ manages to finish training.}. \textit{Normalize + Hierarchical-Sent} is standard \method where we apply hierarchical reward on each sentence within a long-form speech segment. \textit{Normalize + Hierarchical-Doc} is applying the hierarchical reward on the speech segment level if the segment average quality score is below the threshold. 
As shown in the Table \ref{tab:hier}, \method is better than other three methods in quality-latency trade-off. 

\paragraph{Segmentation}

The segmentation method SEGALE we adopt allows for null alignments which accounts for over/under translation. In contrast, mwersegmenter always enforces an alignment by minimizing the word error rate, even when the hypothesis consists of pure gibberish. Combined with the fact that neural metrics are not entirely robust, this may lead to nonsensical hypotheses receiving deceptively high quality scores. As shown in Table~\ref{tab:mwer}, the model trained with mwersegmenter exploits these weaknesses and effectively hacks the reward. In this example, the source speech contains two sentences with corresponding reference translations, while the gibberish hypothesis is segmented by mwersegmenter and each segment attains near-perfect MetricX scores.

\paragraph{Sensitivity to Hyperparameters}
We analyze the sensitivity of \method to three hyperparameters: the quality threshold $q_{\text{thres}}$, the maximum latency $l_{\max}$, and the latency reward weight $\lambda$. For each configuration, we average results over five training runs to reduce variance and obtain more stable estimates. The results are shown in Figure~\ref{fig:ablation_hyperparam_acl6060}. Overall, the default hyperparameter setting used in \method performs best. For the quality threshold $q_\text{thres}$, increasing it to $-3$ leads to substantially higher latency, likely because the model has more difficulty reaching the target quality level and therefore optimizes latency less. Lowering the threshold to $-7$ or $-9$ reduces latency, but also degrades translation quality. For the maximum latency penalty, reducing $l_{\max}$ to 5 hurts translation quality, while increasing it to 20 yields results similar to the default setting of 10. For the latency reward weight, decreasing $\lambda$ to 0.2 results in higher latency, whereas increasing it reduces latency at the cost of worse translation quality.


\section{Conclusion}

In this paper, we propose Hierarchical Policy Optimization (\method) to correct the erroneous behaviors of SFT models trained on imperfect translation trajectories. \method optimizes latency only when the translation quality exceeds a predefined threshold. Experimental results on the ACL 60/60 dev set and RealSI demonstrate that \method outperforms a strong baseline and even surpasses the quality of offline translation. Ablation studies further show that MetricX serves as the most effective quality reward among all tested reward functions, and that the sentence-level hierarchical reward and robust segmentation method are key to the observed improvements. Finally, our case study reveals that \method enhances fluency and adequacy while slightly increasing the risk of omission.

\section*{Limitations}

This paper explores RL-based post-training for SFT models trained on imperfect translation data. However, we consider only a single model architecture InfiniSST, one data synthesis approach with word alignment tool, and three language directions, with English as the sole source language. In addition, our main results and case study reveal that the best-performing reward model, MetricX, is still imperfect. It sometimes favors fluency over accuracy and potentially leads to reward hacking, highlighting the need for more robust quality reward models for SST. 

\bibliography{clean}

\appendix
\section{Appendix}

\subsection{Data Synthesis}
\label{apdx:data_synth}

We first apply the state-of-the-art open-source ASR model parakeet-tdt-0.6b-v2\footnote{\url{https://huggingface.co/nvidia/parakeet-tdt-0.6b-v2}}
 to transcribe English speech with timestamps. Using these timestamps, we group consecutive utterances into long-form segments capped at 67.2 seconds. Each segment is then translated into Chinese, German, and Japanese using Qwen3-32B-AWQ~\citep{qwen3technicalreport}\footnote{\url{https://huggingface.co/Qwen/Qwen3-32B-AWQ}}
, with translation prompts provided in Figure \ref{prompt:fwd_translation}.

To ensure translation quality, we filter translations with Blaser-2.0-QE~\citep{dale-costa-jussa-2024-blaser}\footnote{\url{https://huggingface.co/facebook/blaser-2.0-qe}}
 and MetricX-24-QE~\citep{juraska-etal-2024-metricx}\footnote{\url{https://huggingface.co/google/metricx-24-hybrid-xxl-v2p6-bfloat16}}
, keeping only long-form segments of which all utterances pass both thresholds. Based on preliminary analysis, we set the Blaser-2.0-QE threshold to 3.0 for all three language directions, and the MetricX-24-QE threshold to -4.0 for En–Zh/En–De and -5.0 for En–Ja. After filtering, the resulting dataset contains 1592 hours for En–Zh, 1622 hours for En–De, and 1018 hours for En–Ja. 

\begin{figure}
    \centering
    \includegraphics[width=\linewidth]{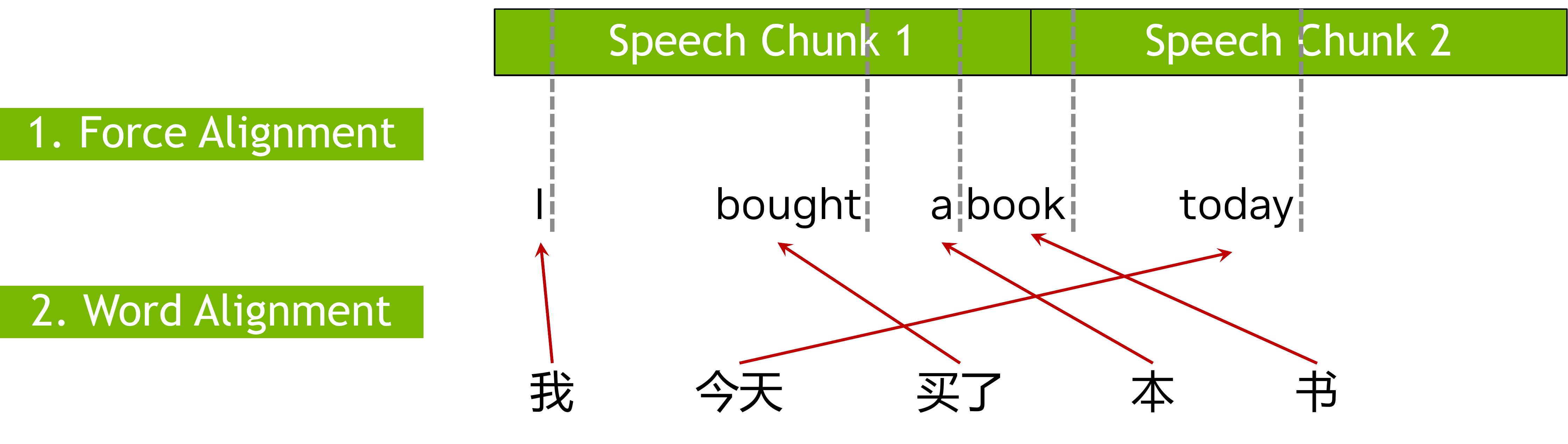}
    \includegraphics[width=\linewidth]{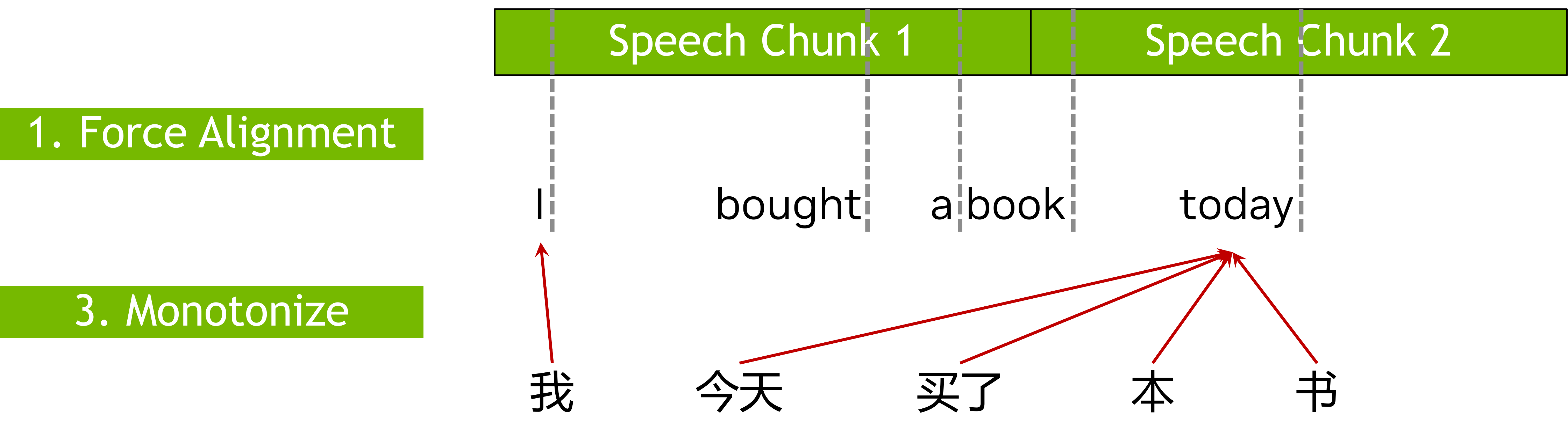}
    \includegraphics[width=\linewidth]{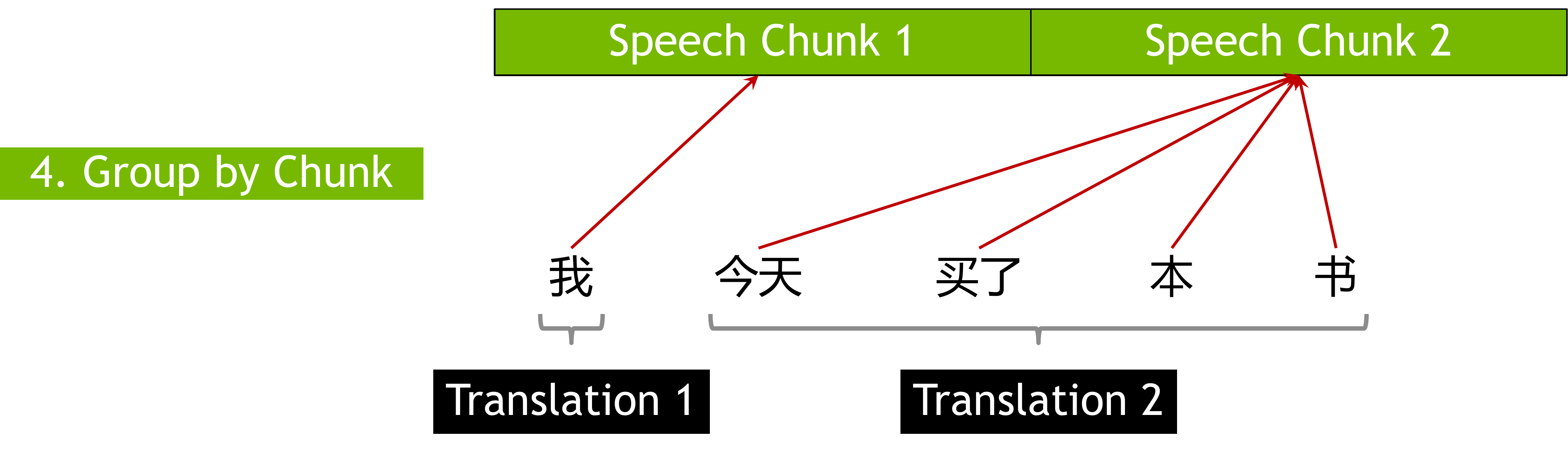}
    \caption{Data Synthesis}
    \label{fig:data-synth}
\end{figure}

Finally, we generate the synthetic trajectories. 
As shown in Figure~\ref{fig:data-synth}, we apply SimAlign~\citep{jalili-sabet-etal-2020-simalign} using LaBSE model~\cite{feng-etal-2022-language} to align words in the source transcript with their counterparts in the target translation. Finally, we enforce monotonicity on the alignment and group target words that correspond to the same speech chunk.

\begin{CJK*}{UTF8}{gkai}
\begin{table*}[t]
\centering
\small
\begin{tabularx}{\textwidth}{lX}
\toprule
\textbf{Category} & \textbf{Example and Observation} \\
\midrule

\textbf{Better Fluency} &
\textbf{Source:} ``Can we start generating the program and executing it before the user even finishes the utterance so that the faster response can be achieved by the system?'' \\
& \textbf{Reference:} 在这次演讲中，我们提出了这样一个问题：我们是否可以在用户完成话语之前就开始生成程序并执行它，以便系统能够实现更快的响应？ \\
& \textbf{SFT:} 在本次演讲中，我们提出一个问题：我们能否在生成程序的同时执行它，在用户尚未完成话语之前就开始呢？因此，系统可以更快地响应。通过系统实现更快的响应。 \\
& \textbf{HPO:} 在本次演讲中，我们提出一个问题：我们能否在用户尚未完成表达之前就开始生成并执行程序？这样可以加快响应速度。 \\
& \textit{Observation:} The HPO translation is more concise and fluent, with smoother phrasing and natural word order. It avoids repetition and better matches the rhythm of natural Chinese speech. \\

\midrule

\textbf{More Accuracy} &
\textbf{Source:} ``To motivate the task, let me begin by defining a counterfactual.'' \\
& \textbf{Reference:} 为了激发这项任务的动机，让我从定义一个反事实开始。 \\
& \textbf{SFT:} 让我先讲一个合同模板的定义。 \\
& \textbf{HPO:} 为了说明这项任务的重要性，我先定义一下反事实。 \\
& \textit{Observation:} The SFT translation is semantically incorrect (“合同模板” / contract template), while HPO correctly captures the meaning of “counterfactual.” This demonstrates improved semantic grounding and disambiguation. \\

\midrule

\textbf{Omission} &
\textbf{Source:} ``Here, at the prefix after Obama we predict confidently that the find person function should be in the program, but the rest may contain errors as they are grayed out.'' \\
& \textbf{Reference:} 在这里，在Obama后面的前缀处，我们自信地预测“查找人员”函数应该在程序中，但其余部分可能包含错误，因为它们显示为灰色。 \\
& \textbf{SFT:} 在奥巴马的前缀处，我们自信地预测查找人物函数应该在程序中。但其余的可能包含错误，如它们是大节点。 \\
& \textbf{HPO:} 在这里，奥巴马这个前缀之后，我们可以自信地预测查找人物函数应该出现在程序中，但其余部分可能包含错误。 \\
& \textit{Observation:} The HPO translation is fluent but omits the causal phrase “因为它们显示为灰色,” showing that MetricX may overemphasize fluency at the expense of completeness. \\

\bottomrule
\end{tabularx}
\caption{Qualitative examples comparing SFT and \method outputs. The HPO model generally improves fluency and accuracy but may occasionally omit minor details.}
\label{tab:qualitative_cases}
\end{table*}
\end{CJK*}

\subsection{Prompt}
\label{apdx:prompt_vip}
Prompt template for forward translation and VIP-LLM are shown in Figure \ref{prompt:fwd_translation} and \ref{prompt:vip-llm}.

\tcbset{
  colback=gray!5,      
  colframe=gray!40,    
  boxrule=0.3pt,
  arc=1pt,
  left=4pt, right=4pt, top=4pt, bottom=4pt,
  fonttitle=\bfseries,
}

\begin{figure*}[t]
\centering
\begin{tcolorbox}[title=Prompt Template for Forward Translation]
\ttfamily
You are given an English document split into lines. Translate each line into Chinese. Do not include any other text.

<begin>

\{source English text\}

<end>

\end{tcolorbox}
\caption{Prompt Template for Forward Translation.}
\label{prompt:fwd_translation}
\end{figure*}

\begin{figure*}[t]
\centering
\begin{tcolorbox}[title=Prompt Template for VIP-LLM]
\ttfamily
[System]
You are a professional translation evaluator.

[User]
Your task is to assess whether a translation segment successfully conveys the semantic content of the original speech according to the following criteria:

1. Key Information Recognition: Identify whether the key information in the source (e.g., proper nouns, keywords, terminologies, or sentence structures) is present in the translation.

2. Correctness Assessment: Determine whether the translation accurately conveys the speaker’s intention, without misinterpretation or contextual errors.

3. Expressiveness Assessment: Evaluate whether the translation is fluent, clear, and intuitive to human readers. It should avoid unnecessary verbosity, ambiguous phrases, or awkward grammar.

Given a source sentence and its translation, answer "Yes" if the translation meets all three criteria and answer "No" otherwise. Only output the answer, no other text.

<begin\_of\_source>

\{source English text\}

<end\_of\_source>

<begin\_of\_translation>

\{translation hypothesis\}

<end\_of\_translation>

\end{tcolorbox}
\caption{Prompt Template for VIP-LLM.}
\label{prompt:vip-llm}
\end{figure*}

\subsection{SFT Training Details}
\label{apdx:sft_train}

We adopt a two-stage SFT procedure. In Stage 1, the LLM is frozen and we train only the speech encoder and the adapter. In Stage 2, we freeze the speech encoder and adapter and fine-tune only the LLM. The global batch size corresponds to \(\sim2.4\) hours of audio. We use Adam with a learning rate of \(1\times10^{-6}\) for Stage 1 and \(4\times10^{-5}\) for Stage 2. Training runs for up to 8k steps in Stage 1 and 2k steps in Stage 2. To increase data diversity, we randomly merge every \(c\) consecutive chunks with \(c \in [1, 12]\).

\subsection{Human Evaluation}
\label{apdx:human_eval}

\begin{figure*}
    \centering
    \includegraphics[width=\linewidth]{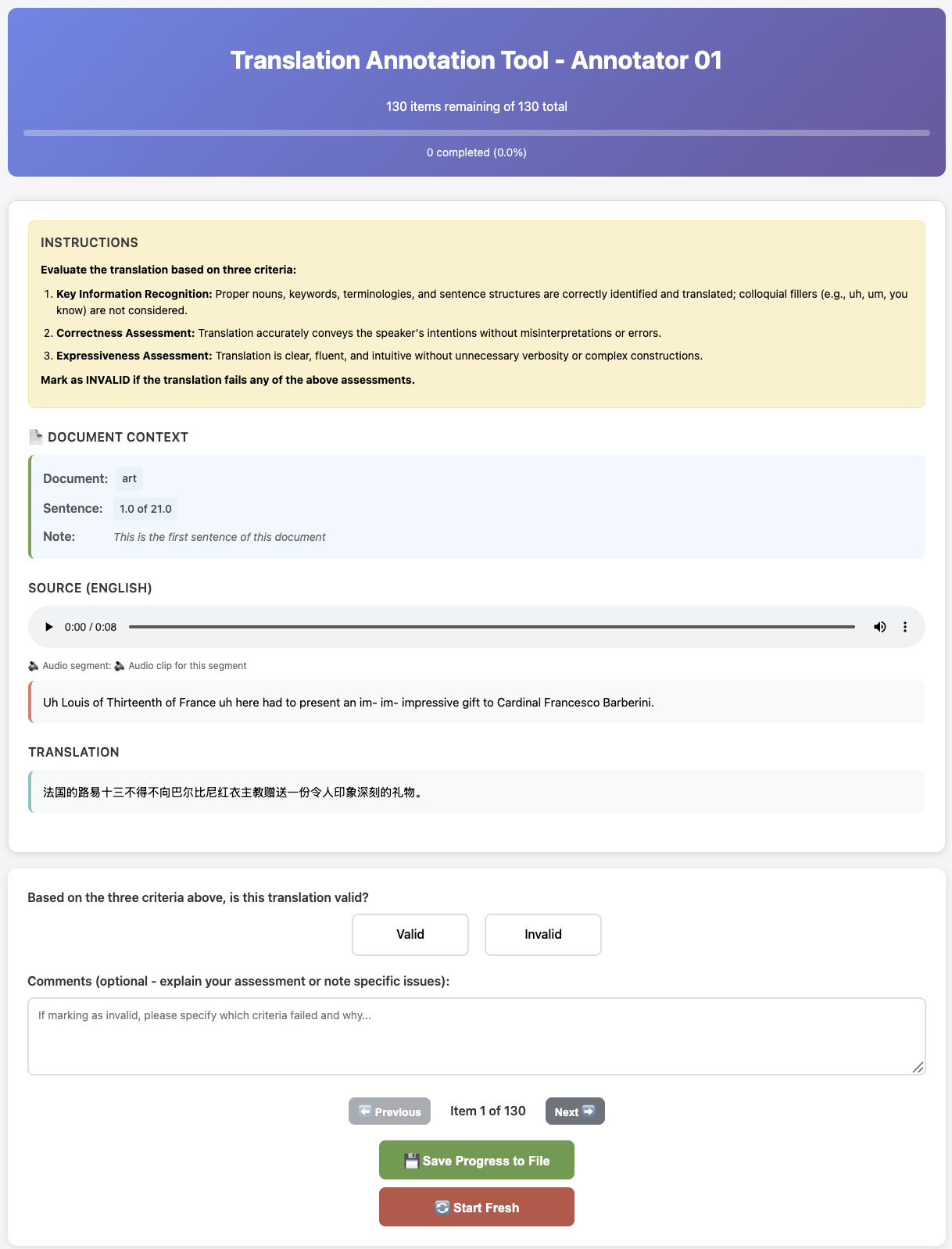}
    \caption{Instruction to human annotators.}
    \label{fig:human_annotator}
\end{figure*}

We provide the screenshot of web application to human annotators in Figure \ref{fig:human_annotator}. We hired human annotators from the university lab and compensated them at the minimum wage rate in the United States.

\subsection{Case Study}
\label{apdx:case_study}

We manually compare 100 SFT model outputs and \method model outputs on the ACL 60/60 dev set to examine how \method influences generation behavior. We identify three major behavioral shifts. First, in 28\% of cases, the \method model waits for the right additional context before generating, resulting in smoother phrasing and more natural word order. 
Second, in an additional 10\% of cases, both outputs are fluent, but the \method model produces translations that are more semantically faithful to the source. 
Interestingly, in 6\% of cases, we observe omissions in the translations of the \method model, likely because the MetricX reward emphasizes fluency and coherence, sometimes causing the model to skip minor details. Typical qualitative examples of each category are provided in Table \ref{tab:qualitative_cases}.

\end{document}